\documentclass[conference]{IEEEtran}
\IEEEoverridecommandlockouts
\usepackage{cite}
\usepackage{amsmath,amssymb,amsfonts}
\usepackage{algorithmic}
\usepackage{graphicx}
\usepackage{textcomp}
\usepackage{xcolor}
\usepackage{enumerate}
\usepackage{enumitem}
\usepackage{booktabs}
\usepackage{multirow}
\usepackage{pifont}
\usepackage{caption}
\usepackage{subcaption}
\usepackage{orcidlink}
\usepackage{hyperref}

\def\BibTeX{{\rm B\kern-.05em{\sc i\kern-.025em b}\kern-.08em
    T\kern-.1667em\lower.7ex\hbox{E}\kern-.125emX}}

\makeatletter
\DeclareRobustCommand\onedot{\futurelet\@let@token\bmv@onedotaux}
\def\bmv@onedotaux{\ifx\@let@token.\else.\null\fi\xspace}
\def\eg{\emph{e.g}.}

 \def\vs{\emph{vs}.}

\makeatother

\begin{document}

\title{Quo Vadis, Anomaly Detection? \\LLMs and VLMs in the Spotlight}


\author{\IEEEauthorblockN{Xi Ding}
\IEEEauthorblockA{\textit{School of Computing} \\
\textit{Australian National University}\\
Canberra, Australia \\
Xi.Ding1@anu.edu.au
}
\and
\IEEEauthorblockN{Lei Wang\thanks{* Corresponding author.} \textsuperscript{$\!\!*$}\orcidlink{0000-0002-8600-7099}}
\IEEEauthorblockA{\textit{School of Computing} \\
\textit{Australian National University}\\
Canberra, Australia \\
Lei.W@anu.edu.au 
}
}

\maketitle

\begin{abstract}

Video anomaly detection (VAD) has witnessed significant advancements through the integration of large language models (LLMs) and vision-language models (VLMs), addressing critical challenges such as interpretability, temporal reasoning, and generalization in dynamic, open-world scenarios. This paper presents an in-depth review of cutting-edge LLM-/VLM-based methods in 2024, focusing on four key aspects: (i) enhancing interpretability through semantic insights and textual explanations, making visual anomalies more understandable; (ii) capturing intricate temporal relationships to detect and localize dynamic anomalies across video frames; (iii) enabling few-shot and zero-shot detection to minimize reliance on large, annotated datasets; and (iv) addressing open-world and class-agnostic anomalies by using semantic understanding and motion features for spatiotemporal coherence. We highlight their potential to redefine the landscape of VAD. Additionally, we explore the synergy between visual and textual modalities offered by LLMs and VLMs, highlighting their combined strengths and proposing future directions to fully exploit the potential in enhancing video anomaly detection.

\end{abstract}


\section{Introduction}
\label{sec:intro}

Video anomaly detection (VAD) is a critical problem with widespread applications in security surveillance, healthcare, autonomous driving, and content moderation \cite{zhao2017spatio, wang2019loss, nguyen2019anomaly, zhu2021video,ren2021deep, samaila2024video,zhuadvancing}. The ability to automatically identify abnormal events or behaviors in video data is essential for real-time intervention, system optimization, and understanding complex dynamics in a variety of domains \cite{nawaratne2019spatiotemporal}. However, traditional approaches \cite{saligrama2012video,zhao2017spatio, leyva2017video,wang2019loss, nguyen2019anomaly, zhu2021video,ren2021deep,hao2022spatiotemporal, sultani2018real, wu2020not, acsintoae2022ubnormal} to VAD face significant challenges due to the dynamic nature of video content, the complexity of detecting anomalies across various contexts, and the difficulty in obtaining labeled data for training robust models \cite{zhuadvancing,10667004}.

Recent advancements in deep learning have introduced powerful models such as large language models (LLMs) and vision-language models (VLMs), which show promising potential in enhancing VAD performance \cite{radford2021learning,zhu2023minigpt, zhou2022learning, ding2024language,ding2024lego}. LLMs and VLMs enable a deeper understanding of both the visual and textual content of videos, offering new possibilities for detecting and explaining anomalies. These models can capture long-range temporal dependencies, understand contextual relationships, and even generate textual descriptions of video content, making them a versatile tool for improving anomaly detection in real-world, open-world scenarios.

Despite these advancements \cite{luo2019video, duong2023deep,ding2024lego}, several challenges remain. First, most existing VAD methods struggle with capturing complex temporal relationships and context, which are often critical for understanding the evolution of anomalies over time \cite{xu2014video}. Second, ensuring interpretability and explainability in anomaly detection is essential for real-world deployment, where transparency in decision-making is crucial \cite{wu2022explainable, ye2024vera}. Third, the availability of labeled training data remains a bottleneck for many VAD systems, especially in open-world scenarios where new and previously unseen anomalies may arise \cite{zhu2022towards, tang2024hawk}. Finally, current methods tend to focus on class-specific anomalies, limiting their ability to generalize to open-world, class-agnostic settings \cite{zhou2019anomalynet, nguyen2019anomaly}.

\begin{figure*}[tbp]
\centering
\begin{subfigure}[t]{0.247\linewidth}
\centering\includegraphics[trim=5.2cm 2.5cm 6.2cm 1.0cm, clip=true, height=2.75cm]{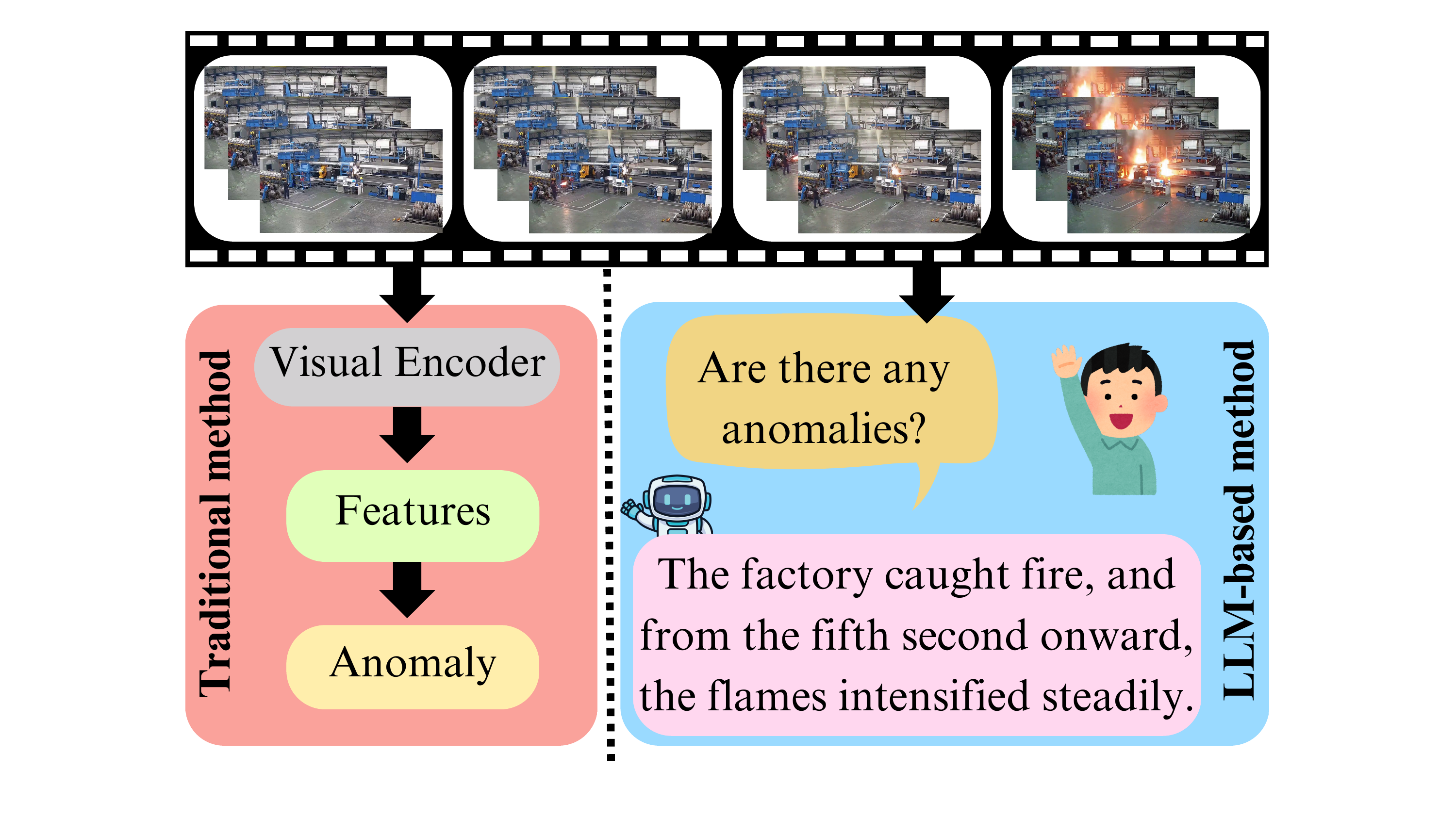}
\caption{Temporal modeling.}\label{fig:temp}
\end{subfigure}
\begin{subfigure}[t]{0.247\linewidth}
\centering\includegraphics[trim=5.6cm 3.5cm 6.0cm 0.8cm, clip=true, height=2.75cm]{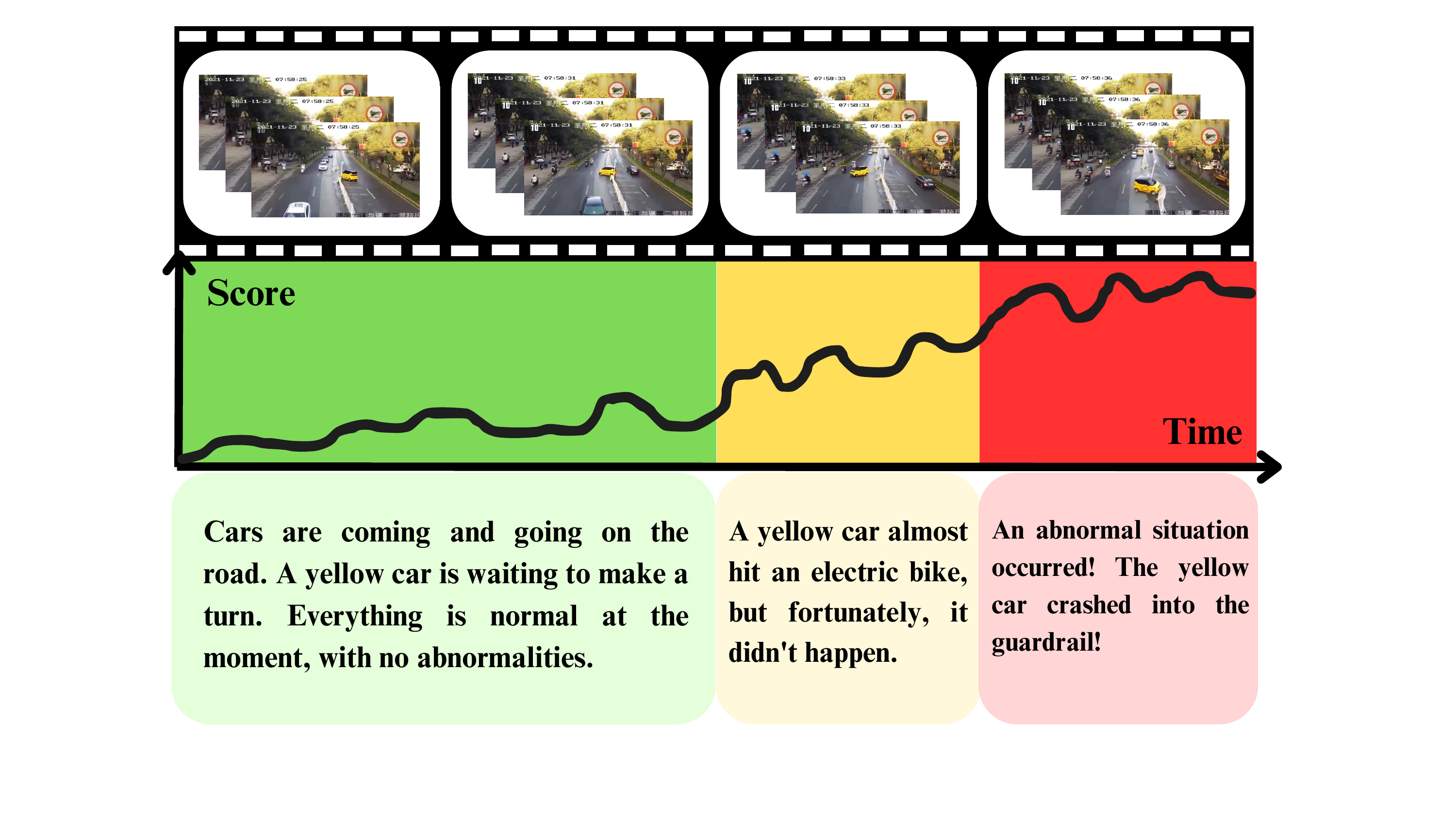}
\caption{Interpretability.}\label{fig:interp}
\end{subfigure}
\begin{subfigure}[t]{0.247\linewidth}
\centering\includegraphics[trim=5.2cm 2.8cm 6.2cm 0.8cm, clip=true, height=2.75cm]{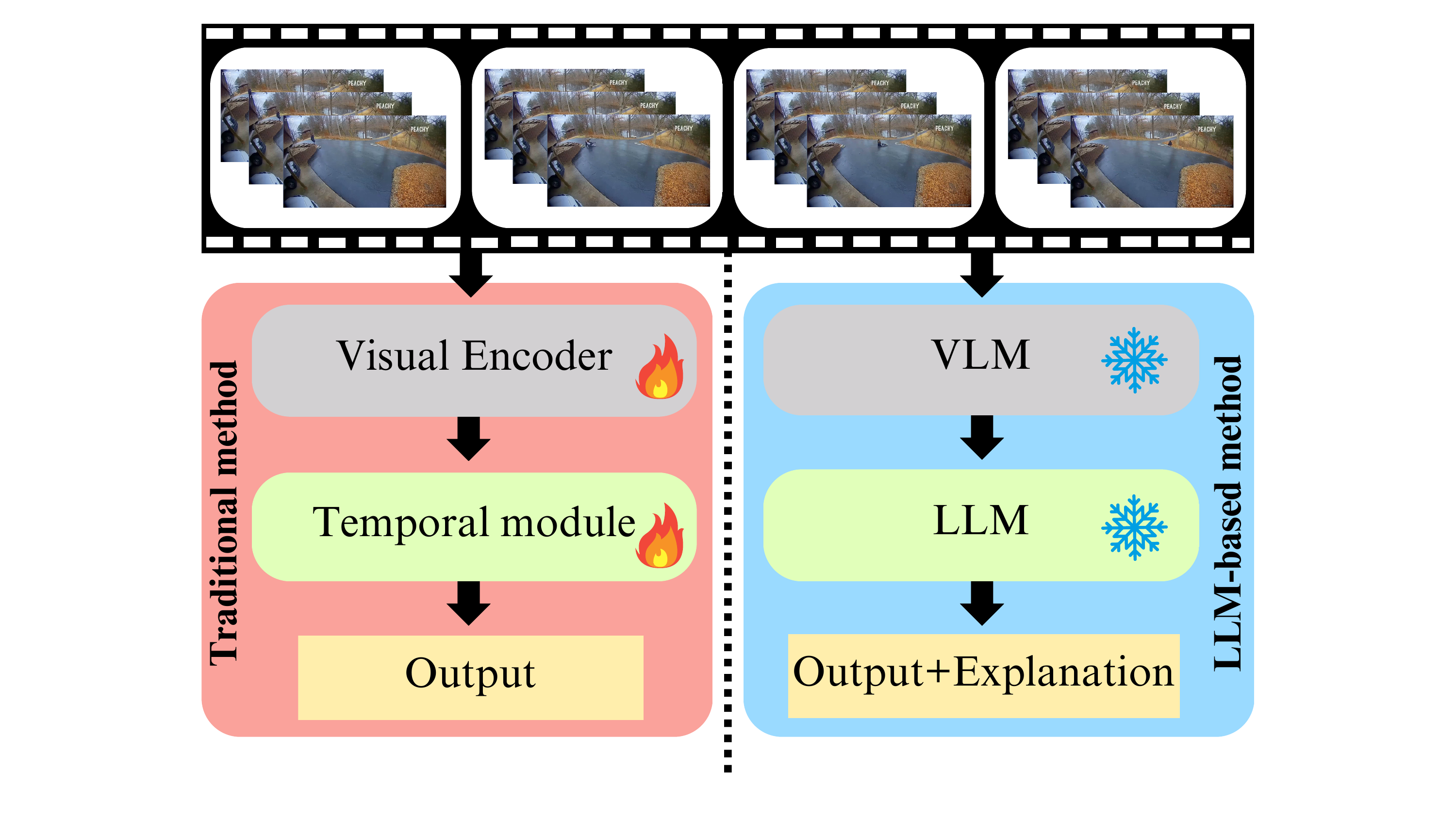}
\caption{Training-free.}\label{fig:train-free}
\end{subfigure}
\begin{subfigure}[t]{0.247\linewidth}
\centering\includegraphics[trim=5.0cm 1.8cm 3.5cm 0.9cm, clip=true, height=2.75cm]{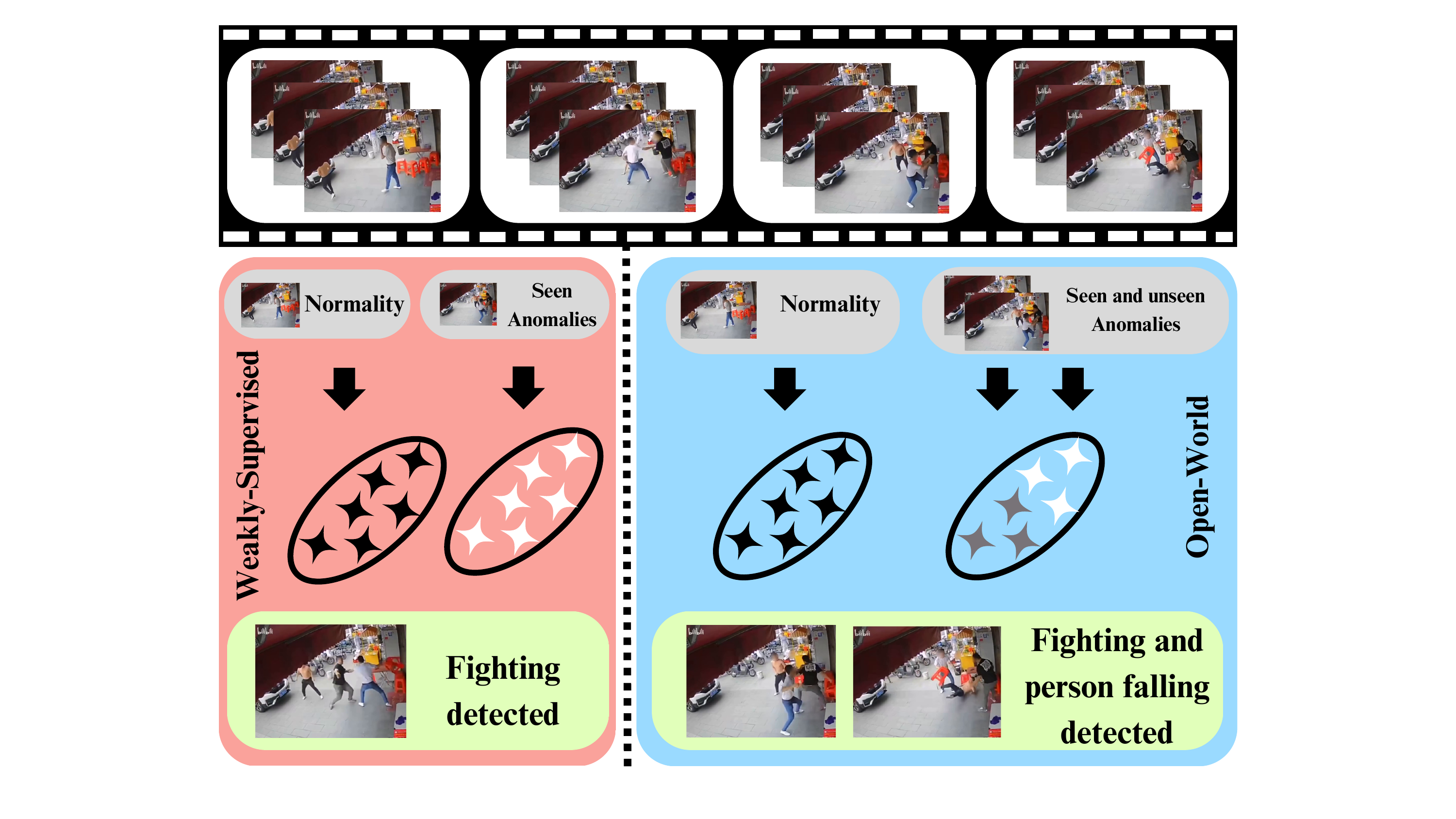}
\caption{Open-world.}\label{fig:open-world}
\end{subfigure}
\caption{We present a systematic evaluation of 13 closely related works from 2024 that use large language models (LLMs) and vision-language models (VLMs) for video anomaly detection (VAD). The analysis is organized around four key perspectives: (a) temporal modeling, (b) interpretability, (c) training-free, and (d) open-world detection, each represented by a subfigure. For each perspective, we highlight the strategies used, key strengths, limitations, and outline promising directions for future research. The video frames used in the analysis are sourced from the MSAD \cite{zhuadvancing} dataset.}
\label{fig:aspect}
\end{figure*}

This work presents a comprehensive review and analysis of recent methods that integrate LLMs and/or VLMs for VAD. To align with current research trends, we examine 13 recently published works from 2024, exploring four critical aspects: temporal and contextual relationships, interpretability and explainability, training-free and few-shot learning approaches, and open-world/class-agnostic anomaly detection (illustrated in Figure \ref{fig:aspect}). We evaluate the strengths and limitations of these approaches, offering valuable insights into how LLM and VLM integration can drive advancements in VAD. The key \textbf{contributions} of this work are as follows:
\renewcommand{\labelenumi}{\roman{enumi}.}
\begin{enumerate}[leftmargin=0.6cm]
\item We identify the latest language model-driven methods, discussing 4 perspectives: temporal modeling, interpretability, training-free learning, and open-world detection.
\item We conduct a comparative analysis of these methods, highlighting their strengths and weaknesses in addressing real-world challenges in VAD.
\item We propose future research directions, emphasizing the integration of temporal context, fine-grained interpretability, and adaptive methods to detect new, unseen anomalies. We suggest that combining training-free approaches with fine-grained semantic supervision and open-world capabilities could enable more robust and scalable VAD solutions.
\end{enumerate}

\section{Related Work}
\label{sec:related}



\textbf{Interpretability and semantic insights.} Interpretability has become a crucial concern in VAD, especially in sensitive or high-stakes applications where it is essential to explain why a particular anomaly was flagged. Early methods \cite{sultani2018real, wu2020not, acsintoae2022ubnormal} often relied on black-box models, which made it difficult to trust their predictions. Recent approaches \cite{lv2024video, zanella2024harnessing, jiang2024vision, zhang2024holmes1, tang2024hawk, wu2024weakly, 10.1145/3664647.3681190} have used semantic insights from LLMs and textual explanations from VLMs to generate intelligible reasoning for detected anomalies. These systems map detected visual anomalies to textual descriptions or semantic cues, making it easier for end-users to understand the nature of the anomalies. While this significantly improves transparency, the challenge remains in balancing the granularity of these explanations with computational efficiency, especially for real-time systems.

\textbf{Temporal reasoning in dynamic anomalies.} Detecting and localizing dynamic anomalies that unfold over time remains one of the central challenges in VAD. Early methods \cite{sultani2018real, wu2020not, acsintoae2022ubnormal} typically analyzed video frames independently, missing out on the temporal relationships that define many anomalies. Recent works \cite{lv2024video, zanella2024harnessing, wu2024open, jiang2024vision, tang2024hawk, yang2024text, zhang2024holmes1} integrating LLMs and VLMs have started to address this gap by modeling long-range dependencies between frames, enabling the detection of anomalies that span across temporal sequences. These models use advanced techniques such as motion and context modeling to improve the capture of temporal dynamics, which are crucial for identifying irregular behaviors in dynamic scenarios. However, scalability and handling noisy or incomplete data remain significant hurdles for these temporal reasoning methods.

\textbf{Few-shot and zero-shot detection.} The scarcity of labeled data is a persistent challenge in VAD, particularly for detecting anomalies in novel, unseen contexts \cite{sultani2018real, wu2020not, acsintoae2022ubnormal}. Few-shot and zero-shot methods, powered by LLMs and VLMs, offer promising solutions by enabling generalization from limited labeled data \cite{zanella2024harnessing, yang2025follow, wu2024open, wu2024weakly, 10.1145/3664647.3681190, ye2024vera}. These models use pre-trained knowledge to recognize anomalies in unseen classes or with minimal training data. Methods that rely on semantic understanding of video content, combined with motion features, make it possible to identify anomalies without the need for large-scale annotated datasets. However, despite the potential, these methods often struggle with complex anomaly types that deviate significantly from the norm, especially in open-world environments where the nature of anomalies is unknown.

\textbf{Open-world and class-agnostic anomalies.} Traditional VAD approaches \cite{sultani2018real, wu2020not, acsintoae2022ubnormal} typically operate in closed-world settings where predefined anomaly classes are assumed. However, real-world applications require models capable of detecting open-world, class-agnostic anomalies, which may involve previously unseen behaviors. Multimodal models \cite{wu2024open, zanella2024harnessing, wu2024weakly, zhang2024holmes2, ntelopoulos2024callm} that combine semantic and motion reasoning are making strides in addressing these challenges by detecting anomalies without prior knowledge of the specific class. These systems are more robust in open-world settings, where they can detect unexpected anomalies, but issues related to scalability and dynamic adaptation remain unresolved, particularly when new types of anomalies appear over time.

\textbf{Motivation and key differences.} While existing methods have made significant strides in one or more of these areas \cite{zhuadvancing,ding2024language,ding2024lego}, the integration of LLMs and VLMs offers a holistic approach to the challenges of video anomaly detection. Unlike previous works that tend to focus on isolated aspects of VAD (\eg, temporal reasoning, interpretability, or class-specific detection), this paper emphasizes the synergy between visual and textual modalities to address all key challenges simultaneously. By focusing on semantic insights and motion features, this review highlights how multimodal models can provide a more comprehensive solution to video anomaly detection. Moreover, by exploring few-shot and zero-shot capabilities, this paper proposes a shift toward more generalizable systems that can perform well even with minimal training data.

\section{Insights on Recent Advances}
\label{sec:discussion}

\begin{table*}[tbp]
	\begin{minipage}{0.65\linewidth}
		\caption{Comparison of different sampling strategies for temporal reasoning.}
		\label{tab:sampling-strategies}
		\centering
\begin{tabular}{lccclc}
\toprule
\textbf{Sampling} & \textbf{Interval} & \textbf{Frame count} & \textbf{Redundancy} & \textbf{Target use case}       & \textbf{Cost} \\ \midrule
\textbf{Uniform}           & Fixed                   & Medium                  & Medium                          & Global trend               & High                               \\
\textbf{Random}             & Random                  & Medium                  & Low                             & Data augmentation          & High                               \\
\textbf{Key frame}         & Adaptive                & Low to Med.           & Low                             & Key event extraction       & Medium                             \\
\textbf{Dense}           & One                     & High                    & High                            & Fine-grained modeling      & Low                                \\
\textbf{Sliding window}     & Adaptive                & Medium                  & Medium                          & Local temporal details     & Medium                             \\
\textbf{Adaptive}           & Dynamic                 & High                    & Low                             & Comprehensive modeling     & Medium                             \\ \bottomrule
\end{tabular}
\end{minipage}\hfill
\begin{minipage}{0.3\linewidth}
\centering
\includegraphics[width=0.9\linewidth, trim=2.8cm 2.8cm 2.8cm 2.8cm, clip=true]{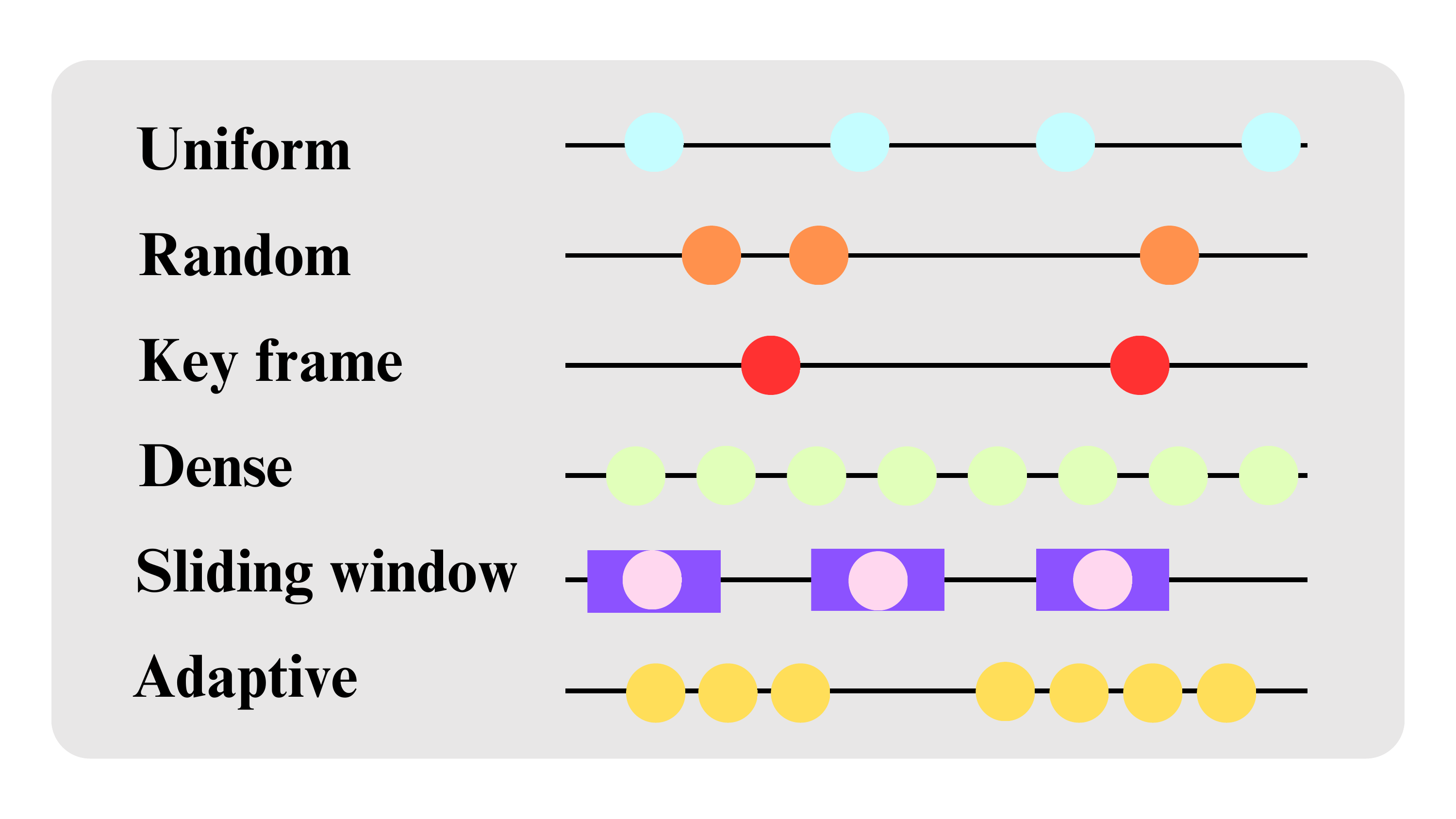}
  \captionof{figure}{Various sampling strategies.}
  \label{fig:samp-vis}
\end{minipage}
\end{table*}

We offer a thorough analysis in VAD in 2024, with a focus on the integration of LLMs and VLMs. The methods reviewed include:  \texttt{VADor} \cite{lv2024video}, \texttt{OVVAD} \cite{wu2024open}, \texttt{LAVAD} \cite{zanella2024harnessing}, \texttt{TPWNG} \cite{yang2024text}, \texttt{CALLM} \cite{ntelopoulos2024callm}, \texttt{Holmes-VAD} \cite{zhang2024holmes1}, \texttt{HAWK} \cite{tang2024hawk}, \texttt{VLAVAD} \cite{jiang2024vision}, \texttt{ALFA} \cite{10.1145/3664647.3681190}, \texttt{AnomalyRuler} \cite{yang2025follow}, \texttt{STPrompt} \cite{wu2024weakly}, \texttt{Holmes-VAU} \cite{zhang2024holmes2}, and \texttt{VERA} \cite{ye2024vera}.
We structure our discussion around four key perspectives (see Figure \ref{fig:aspect}) addressed by these recent advancements: temporal and contextual relationships, interpretability and explainability, training-free and few-shot learning approaches, and open-world/class-agnostic anomaly detection. For each perspective, we highlight the strategies used, evaluate the strengths and limitations of the methods, and suggest potential directions for future research.

\subsection{Temporal Modeling and Context}

Temporal modeling is fundamental to video anomaly detection (VAD), as anomalies are often characterized by deviations in temporal patterns. The primary challenge lies in capturing intricate temporal dynamics while maintaining computational efficiency and scalability. Recent methods address these challenges with innovative modules and the integration of contextual reasoning \cite{lv2024video, zanella2024harnessing, wu2024open, jiang2024vision, tang2024hawk, yang2024text, zhang2024holmes1}.

\texttt{VADor} \cite{lv2024video} introduces a Long-Term Context (LTC) module to address the limitations of open-sourced video LLMs in handling long-range context, effectively capturing temporal dynamics. However, scalability remains an issue for longer or more complex videos. \texttt{LAVAD} \cite{zanella2024harnessing} uses a sliding window over frame-level captions to aggregate temporal information, achieving reasonable performance in structured scenarios but faltering with noisy or incomplete captions.
\texttt{OVVAD} \cite{wu2024open} uses a graph convolutional network (GCN) as a temporal adapter, bridging frozen CLIP encoders with sequential data for effective temporal reasoning without extensive retraining. However, it struggles to fully exploit fine-grained temporal cues. \texttt{VLAVAD} \cite{jiang2024vision} integrates semantic inconsistencies with temporal information through a Sequence State Space Module (S3M), improving anomaly detection in unsupervised settings but facing scalability challenges due to high-dimensional state representations.
Motion-centric approaches, such as \texttt{HAWK} \cite{tang2024hawk}, use motion-to-language mappings to connect dynamic patterns with textual descriptions, enhancing interpretability and precision in motion anomalies. Similarly, \texttt{TPWNG} \cite{yang2024text} adapts to varying video durations using self-learning modules, excelling in weakly supervised settings. Finally, \texttt{Holmes-VAD} \cite{zhang2024holmes1} combines a lightweight temporal sampler with multimodal analysis, effectively identifying and explaining anomalies in complex scenarios.

These approaches showcase a diverse range of temporal modeling strategies. While \texttt{VADor} and \texttt{OVVAD} focus on predefined modules for long-term context, \texttt{HAWK} and \texttt{Holmes-VAD} emphasize motion dynamics and adaptive sampling. Future work could combine motion-based features \cite{wang2024flow, wangtaylor, chen2024motion, chen2024spatial, raj2024tracknetv4} with advanced context-aware modules to address scalability and efficiency challenges in real-time anomaly detection.

\subsection{Interpretability and Transparency}

Interpretability is increasingly recognized as a critical factor in VAD systems, particularly for deployment in sensitive and high-stakes environments. Methods in this category focus on generating semantic and multimodal insights, making anomaly detection systems more comprehensible to end-users \cite{lv2024video, zanella2024harnessing, jiang2024vision, zhang2024holmes1, tang2024hawk, wu2024weakly, 10.1145/3664647.3681190}.

\texttt{VADor} \cite{lv2024video} enhances interpretability by fine-tuning Video-LLaMA’s projection layer, blending anomaly detection with semantic reasoning. However, its reliance on instruction-tuned data limits adaptability to diverse anomaly types. \texttt{LAVAD} \cite{zanella2024harnessing} increases transparency through scene descriptions, though noisy captions undermine reliability.
In contrast, \texttt{VLAVAD} \cite{jiang2024vision} simplifies semantic mappings to improve interpretability in unsupervised settings, sacrificing fine-grained detail for reduced complexity. \texttt{Holmes-VAD} \cite{zhang2024holmes1} uses multimodal instruction tuning and temporal supervision to generate context-rich explanations of anomalies. \texttt{HAWK} \cite{tang2024hawk} integrates motion-based reasoning via interactive visual-language models, enhancing interpretability. Similarly, \texttt{STPrompt} \cite{wu2024weakly} aligns spatiotemporal regions with learned prompts, reducing background noise and improving spatial localization. \texttt{ALFA} \cite{10.1145/3664647.3681190} emphasizes pixel-level precision using image-text alignment but requires additional fine-tuning for effective generalization.

The emphasis on semantic and multimodal strategies marks a promising shift toward transparent VAD systems. While \texttt{Holmes-VAD} excels in providing contextual explanations, \texttt{ALFA} offers granular insights. Future research should balance granularity, semantic generalization, and computational efficiency to develop robust, interpretable VAD systems.

\subsection{Training-Free and Few-Shot Detection}

The scarcity of annotated datasets presents a significant challenge for VAD, especially in open-world scenarios. Training-free and few-shot approaches use pre-trained models and minimal annotations to facilitate anomaly detection in data-scarce environments \cite{zanella2024harnessing, yang2025follow, wu2024open, wu2024weakly, 10.1145/3664647.3681190, ye2024vera}.

\texttt{LAVAD} \cite{zanella2024harnessing} bypasses dataset-specific training by using pre-trained LLMs and VLMs for temporal aggregation. While adaptable, its lack of specialization hinders performance with complex anomaly types. \texttt{AnomalyRuler} \cite{yang2025follow} excels in static few-shot scenarios using rule-based reasoning with minimal normal samples but struggles with dynamic anomalies.
\texttt{OVVAD} \cite{wu2024open} decouples anomaly detection from classification, enabling robust detection of unseen anomalies but lacking temporal depth. \texttt{STPrompt} \cite{wu2024weakly} aligns spatiotemporal prompts to localize anomalies under weak supervision, performing well in straightforward cases but faltering with nuanced patterns. \texttt{ALFA} \cite{10.1145/3664647.3681190} dynamically adapts prompts at runtime for fine-grained detection, and \texttt{VERA} \cite{ye2024vera} introduces verbalized learning to enable training-free anomaly detection without modifying model parameters.

Combining the adaptability of \texttt{VERA} with the fine-grained capabilities of \texttt{ALFA}, alongside temporal reasoning as seen in \texttt{OVVAD}, could provide a pathway to more robust solutions for open-world anomaly detection.

\begin{table*}[tbp]
\centering
\caption{
Comparison of recent methods in video anomaly detection (VAD). We compare recent approaches in VAD, highlighting key aspects such as interpretability, temporal modeling, few-shot learning, and open-world detection. Performance is evaluated across six benchmark datasets: UCSD Ped2 (Ped2) \cite{wang2010anomaly}, CUHK Avenue (CUHK)\cite{lu2013abnormal}, ShanghaiTech (ShT)\cite{luo2017revisit}, UCF-Crime (UCF)\cite{sultani2018real}, XD-Violence (XD)\cite{wu2020not}, and UBnormal (UB)\cite{acsintoae2022ubnormal}. Datasets evaluated using Area Under the Curve (AUC) include Ped2, CUHK, ShT, UCF, and UB, while the XD dataset is evaluated using Average Precision (AP).
}
\resizebox{\textwidth}{!}{
\begin{tabular}{llcccccccccccc}
\toprule
\multirow{2}{*}{\textbf{Method}} & \multirow{2}{*}{\textbf{LLM/VLM}} & \multicolumn{4}{c}{\textbf{Property}} & & \multicolumn{6}{c}{\textbf{Dataset}}\\
\cline{3-6}
\cline{8-13}
\addlinespace[0.8ex]  
& & \textbf{Interpret.} & \textbf{Temporal} & \textbf{Few-shot} & \textbf{Open-world} & & \textbf{Ped2} & \textbf{CUHK} & \textbf{ShT} & \textbf{UCF} & \textbf{XD} & \textbf{UB} \\
\midrule
\textbf{VLAVAD}\cite{jiang2024vision} & Fine-tuning & \ding{51} & \ding{51} &  &  & & 99.0 & 87.6 & 87.2 & -- & -- & --\\
\textbf{VADor}\cite{lv2024video} & Fine-tuning & \ding{51} & \ding{51} &  &  & & -- & -- & -- & 88.1 & -- & -- \\
\textbf{OVVAD}\cite{wu2024open} & Fine-tuning &  & \ding{51} &  & \ding{51} &&  -- & -- & -- & 86.4 & 66.5 & 62.9\\
\textbf{LAVAD}\cite{zanella2024harnessing} & Training-free & \ding{51} & \ding{51} & \ding{51} & & &-- & -- & -- & 80.3 & 62.0 & --\\
\textbf{TPWNG}\cite{yang2024text} & Fine-tuning &  & \ding{51} &  &  && -- & -- & -- & 87.8 & 83.7 & --\\
\textbf{Holmes-VAD}\cite{zhang2024holmes1} & Fine-tuning  & \ding{51} & \ding{51} &  &  & & -- & -- & -- & 89.5 & 90.7 & --\\
\textbf{AnomalyRuler}\cite{yang2025follow} & Fine-tuning  &  &  & \ding{51} &  & & 97.9 & 89.7 & 85.2 & -- & -- & 71.9 \\
\textbf{STPrompt}\cite{wu2024weakly} & Fine-tuning  & \ding{51} & \ding{51} &  &  & & -- & -- & 97.8 & 88.1 & -- & 64.0\\
\textbf{Holmes-VAU}\cite{zhang2024holmes2} & Fine-tuning  &  &  \ding{51} &  & \ding{51} & & -- & -- & -- & 89.0 & 87.7 & --\\
\textbf{VERA}\cite{ye2024vera} & Training-free & \ding{51} &  &  &  & & -- & -- & -- & 86.6 & 88.2 & --\\
\bottomrule
\end{tabular}}
\label{tab:exist-results}
\end{table*}

\subsection{Open-World and Class-Agnostic Detection}

Real-world applications demand VAD systems capable of detecting unseen anomalies and adapting to unpredictable scenarios. Open-world and class-agnostic approaches aim to address these challenges \cite{wu2024open, zanella2024harnessing, wu2024weakly, zhang2024holmes2, ntelopoulos2024callm}.

\texttt{OVVAD} \cite{wu2024open} uses a dual-task strategy for both class-agnostic and class-specific detection, though its temporal modeling could be enhanced. \texttt{LAVAD} \cite{zanella2024harnessing} uses textual descriptions for anomaly scoring but is limited by noisy captions. \texttt{STPrompt} \cite{wu2024weakly} excels in weak supervision, localizing anomalies effectively, though its robustness against complex patterns is limited.
\texttt{Holmes-VAU} \cite{zhang2024holmes2} uses hierarchical annotations for broader coverage, while \texttt{CALLM} \cite{ntelopoulos2024callm} innovates with pseudo-labeling using multimodal features, though further validation in dynamic contexts is needed.

Integrating the hierarchical annotation approach of \texttt{Holmes-VAU} with the multimodal innovation of \texttt{CALLM} could address real-world complexities. Further advancements in temporal and textual reasoning frameworks are essential to enhance detection reliability in open-world scenarios.

\section{Analysis and Discussion}
\label{se:result}

\textbf{Frame sampling strategies.} Frame sampling strategies play a pivotal role in balancing temporal resolution, computational efficiency, and overall model performance. Table~\ref{tab:sampling-strategies} summarizes common strategies, while Figure~\ref{fig:samp-vis} visually compares them. 
Dense sampling offers the highest temporal granularity, essential for detecting nuanced, rapid anomalies such as sudden behavioral changes or fleeting events. 
However, the redundancy of densely sampled frames increases computational costs, making this strategy less practical for large-scale or real-time applications.
Uniform sampling, used in methods like \texttt{VERA}, provides a simpler alternative by sampling frames at fixed intervals. This approach balances computational overhead and temporal coverage but often misses critical local temporal patterns. Similarly, random sampling, such as in \texttt{VADor}, introduces variability, augmenting training data by exposing the model to diverse temporal patterns. However, this strategy risks overlooking key anomaly-defining frames, reducing its effectiveness in scenarios requiring precise temporal modeling.
Adaptive sampling, used in \texttt{Holmes-VAD} and \texttt{Holmes-VAU}, dynamically focuses on regions of interest in time. This method prioritizes frames likely to contain anomalies, enabling both fine-grained detection and computational efficiency. Adaptive strategies strike an optimal balance, excelling in scenarios where anomalies are temporally sparse or context-dependent. Nonetheless, their reliance on additional heuristic or learning mechanisms introduces moderate costs.

The choice of sampling strategy should align with the nature of anomalies and the operational constraints. For global trends, uniform or random sampling suffices, while dense or adaptive sampling is indispensable for fine-grained, time-sensitive detection tasks. Integrating adaptive mechanisms into training-free frameworks, as a future direction, could enhance both scalability and precision in VAD systems.

\textbf{Fine-tuning \vs ~training-free approaches.}
Fine-tuning-based methods dominate in datasets requiring detailed temporal reasoning. For example, \texttt{Holmes-VAD} achieves 89.5\% on \textit{UCF-Crime} and 90.7\% on \textit{XD-Violence}, thanks to anomaly-aware fine-tuning that captures temporal and semantic patterns. \texttt{STPrompt}, using spatio-temporal prompts, performs exceptionally on \textit{ShanghaiTech} (97.8\%) but requires retraining for each dataset, limiting scalability.
Training-free methods like \texttt{LAVAD} and \texttt{VERA} excel in scalability, avoiding the overhead of retraining while maintaining competitive performance. \texttt{VERA} achieves 88.2\% on \textit{XD-Violence}, demonstrating its adaptability to new scenarios. However, they may struggle with complex temporal dynamics, \eg, \texttt{LAVAD}'s lower performance on \textit{XD-Violence} (62.0\%) and \textit{UCF-Crime} (80.3\%).

A hybrid approach combining training-free scalability with fine-tuning precision could address these limitations. For instance, integrating temporal sampling techniques from fine-tuning-based methods into training-free frameworks may enhance their temporal reasoning without compromising scalability.
Future research should also explore few-shot learning and open-vocabulary techniques to bridge gaps in generalization and adaptability, as demonstrated by \texttt{Holmes-VAU}'s promising results (89.0\% on \textit{UCF-Crime}). This direction can enable systems to handle emerging anomalies with minimal retraining while maintaining high accuracy.

\textbf{Quantitative evaluation and comparative analysis.} Table~\ref{tab:exist-results} highlights the performance and properties of recent VAD methods across benchmark datasets.
Among the methods evaluated, \texttt{VLAVAD}, \texttt{VADor}, \texttt{Holmes-VAD}, and \texttt{STPrompt} stand out for their high interpretability and temporal modeling, though they perform differently across benchmark datasets. \texttt{VLAVAD} excels in capturing fine-grained temporal features through fine-tuning and is highly effective on datasets such as \textit{UCSD Ped2} (99.0\%), but it lacks adaptability to open-world anomalies. In contrast, \texttt{LAVAD} offers interpretability with semantic explanations, but its performance on datasets like \textit{UCF-Crime} (80.3\%) and \textit{XD-Violence} (62.0\%) is limited due to its insufficient handling of temporal dynamics. This contrast highlights the importance of balancing interpretability with strong temporal modeling for real-world anomaly detection.

In terms of temporal modeling, methods such as \texttt{Holmes-VAD} and \texttt{Holmes-VAU} are more successful in addressing the temporal dependencies inherent in video anomaly detection. \texttt{LAVAD} offers a training-free solution with temporal aggregation, but it struggles to compete with methods like \texttt{TPWNG} that use spatio-temporal prompt learning. 
Despite \texttt{AnomalyRuler} achieving solid performance on the \textit{ShanghaiTech} (85.2\%) dataset, it lags behind \texttt{STPrompt} (97.2\%), demonstrating that \texttt{STPrompt}'s ability to adapt to temporal dynamics in video sequences provides a significant advantage. However, while \texttt{STPrompt} shows strong performance in time-sensitive anomaly detection, its dependence on fine-tuning limits its scalability and applicability to unseen anomaly types, which is a key drawback (\eg, 64.0\% on \textit{UBnormal}).

Few-shot and open-world detection capabilities are critical for handling emerging or previously unseen anomalies, and methods such as \texttt{OVVAD} and \texttt{AnomalyRuler} perform well in this regard. \texttt{OVVAD} shows the ability to detect both seen and unseen anomalies, especially with its open-vocabulary approach and class-agnostic detection. However, its performance is suboptimal in scenarios requiring temporal modeling, as seen with its results on \textit{XD-Violence} (66.5\%). On the other hand, \texttt{AnomalyRuler} achieves strong performance on both \textit{UCSD Ped2} (97.9\%) and \textit{CUHK Avenue} (89.7\%), showcasing its robustness. Its rule-based approach, however, may struggle with more complex, dynamic anomalies, suggesting that while \texttt{AnomalyRuler} is effective in controlled settings, it may need further refinement for broader use cases.

Lastly, the \texttt{Holmes-VAD} and \texttt{STPrompt} methods excel in terms of interpretability, temporal modeling, and adaptability. \texttt{Holmes-VAD} stands out as one of the top performers, especially on the \textit{UCF-Crime} (89.5\%) and \textit{XD-Violence} (90.7\%) dataset, thanks to its combination of anomaly-aware supervision and fine-tuning, which allows it to capture both temporal and semantic features effectively. Similarly, \texttt{STPrompt} uses spatio-temporal prompt learning and fine-tuning to achieve excellent results on datasets like \textit{ShanghaiTech} (97.8\%) and \textit{UCF-Crime} (88.1\%). However, both methods are limited by their reliance on fine-tuning, which reduces their generalization ability across different anomaly types and datasets.

The results indicate that a multi-faceted approach is needed to optimize VAD systems. Methods like \texttt{Holmes-VAD} and \texttt{STPrompt} show that combining fine-tuned temporal and semantic modeling with interpretability and adaptability to new anomalies can lead to high performance across multiple datasets. However, the challenges of scalability, the need for robust temporal models, and handling noisy captions or incomplete annotations remain significant hurdles. The combination of training-free solutions with fine-tuning, as demonstrated in \texttt{LAVAD}, could provide a more versatile framework for open-world anomaly detection.

\section{Conclusion}
\label{sec:concl}

This work explores the integration of large language models (LLMs) and vision-language models (VLMs) in video anomaly detection (VAD), focusing on key challenges such as temporal modeling, interpretability, few-shot learning, and open-world anomaly detection. We examine how recent advances seek to address these challenges, highlighting both the strengths and limitations of current methods.
Our analysis emphasizes the need for more robust temporal modeling to capture complex dependencies within video data, as well as the importance of fine-grained interpretability to better understand anomaly detection decisions. Additionally, we recognize the potential of training-free and few-shot learning methods, which show promise for improving scalability and adaptability in scenarios with limited supervision or previously unseen anomalies.
We propose that future VAD systems could benefit from combining these approaches, such as improving temporal consistency, aligning semantic features, and incorporating adaptive learning strategies. This work lays the foundation for advancing VAD by refining these models, enhancing their scalability, and addressing the complexities inherent in dynamic video data.

\section*{Acknowledgment}
Xi Ding, a Research Assistant with the Temporal Intelligence and Motion Extraction (TIME) Lab at ANU, contributed to this work. TIME Lab is a dynamic research team comprising master’s and honours students focused on advancing video processing and motion analysis. This research was conducted under the supervision of Lei Wang.

\bibliographystyle{IEEEbib}
\bibliography{icme2025references}

\end{document}